\documentclass[conference,a4paper]{IEEEtran}
\IEEEoverridecommandlockouts
\usepackage{cite}
\usepackage{amsmath,amssymb,amsfonts}
\usepackage{algorithmic}
\usepackage{siunitx}
\usepackage[pdftex]{graphicx}
\graphicspath{{..images/}{../images/}}
\DeclareGraphicsExtensions{.pdf,.jpeg,.png,.svg}
\usepackage{textcomp}
\usepackage{xcolor}
\def\BibTeX{{\rm B\kern-.05em{\sc i\kern-.025em b}\kern-.08em
    T\kern-.1667em\lower.7ex\hbox{E}\kern-.125emX}}

\begin{document}

\title{Detecting and Classifying Bio-Inspired Artificial Landmarks Using In-Air 3D Sonar}

 \author{\IEEEauthorblockN{Maarten de Backer\IEEEauthorrefmark{1}, Wouter Jansen\IEEEauthorrefmark{1}\IEEEauthorrefmark{2}\IEEEauthorrefmark{3}, Dennis  Laurijssen\IEEEauthorrefmark{1}\IEEEauthorrefmark{2}, Ralph Simon\IEEEauthorrefmark{4}\IEEEauthorrefmark{1},
 Walter Daems\IEEEauthorrefmark{1}\IEEEauthorrefmark{2},
 Jan Steckel\IEEEauthorrefmark{1}\IEEEauthorrefmark{2}}
 \IEEEauthorblockA{\IEEEauthorrefmark{1}FTI Cosys-Lab, University of Antwerp, Antwerp, Belgium}
 \IEEEauthorblockA{\IEEEauthorrefmark{2}Flanders Make Strategic Research Centre, Lommel, Belgium}
 \IEEEauthorblockA{\IEEEauthorrefmark{4}Behavioural Ecology and Conservation Lab, Nuremberg Zoo, Nuremberg, Germany\\
 \IEEEauthorrefmark{3}wouter.jansen@uantwerpen.be}}

\maketitle

\begin{abstract}
    Various autonomous applications rely on recognizing specific known landmarks in their environment. For example, Simultaneous Localization And Mapping (SLAM) is an important technique that lays the foundation for many common tasks, such as navigation and long-term object tracking. This entails building a map on the go based on sensory inputs which are prone to accumulating errors. Recognizing landmarks in the environment plays a vital role in correcting these errors and further improving the accuracy of SLAM. The most popular choice of sensors for conducting SLAM today is optical sensors such as cameras or LiDAR sensors. These can use landmarks such as QR codes as a prerequisite. However, such sensors become unreliable in certain conditions, e.g., foggy, dusty, reflective, or glass-rich environments. Sonar has proven to be a viable alternative to manage such situations better. However, acoustic sensors also require a different type of landmark. In this paper, we put forward a method to detect the presence of bio-mimetic acoustic landmarks using support vector machines trained on the frequency bands of the reflecting acoustic echoes using an embedded real-time imaging sonar.
\end{abstract}

\begin{IEEEkeywords}
Acoustic sensors, Sonar, Robot sensing systems, Simultaneous localization and mapping
\end{IEEEkeywords}

\section{Introduction}
    % Intro intro
    Simultaneous Localization And Mapping (SLAM) is essential in navigating known and unknown environments. In the latter case, SLAM proves especially vital without external referencing systems such as the Global Positioning System (GPS). These include complex environments such as cave and tunnel systems seen during search and rescue missions or traversed by mining vehicles. Lastly, it provides the basis for essential tasks such as long-term object tracking, path planning, and collision avoidance \cite{Esrafilian2017, Bescos2021}. \\       
    % Sensors
    Optical sensors, such as cameras and laser scanners (LiDAR), are the preferred input choice for SLAM systems. However, under certain conditions, these sensors will drop in reliability, e.g., foggy, dusty, reflective, or glass-rich environments \cite{Debeunne2020}. \\
    In such situations, in-air sonar can provide a reliable backup, as these factors do not hinder sonar operation using the acoustic ultrasound spectrum. Furthermore, sonar has been proven to be a viable primary sensor for autonomous systems such as SLAM \cite{Batslam}. \\Additional benefits include low cost and the ability to deploy transducer and microphone arrays for a wide field of view and more complex signal processing, such as beamforming \cite{Chen2006, michel2006history}. This last option, in particular, is interesting since it provides increased directionality and improves sensing range.\\    
    % Landmarks
    A priori landmarks improve the reliability of SLAM by correcting errors and increasing the accuracy of the SLAM process \cite{thrun2005, Sefati2017} as odometry data fails to produce an acceptable map, and sensory input is vulnerable to ambiguous measurements, including when using in-air sonar \cite{Batslam}. Landmark recognition also provides the foundation of loop-closure algorithms\cite{Grisetti2010}. \\
    Landmarks can also be ideal for other autonomous tasks, such as specific behavior activation. For example, placing them next to a door to be opened, a mobile robot's charging dock to park, or recognizing a box to be picked up. Cameras are known to use landmarks such as fiducial markers \cite{kalaitzakis2021fiducial}.\\    
    % Conclusion
    Objects can be recognizable by sonar based on their acoustic reflection patterns, which nature reflects in the evolution of bat-pollinated plants\cite{test}. Like sonar, bats produce ultrasound chirps and extract information about the environment based on their reflections. They are competent creatures, as they can differentiate multiple interfering echos, identify the type of vegetation that produced the echo, and resolve surface structures down to \SI{100}{\micro\meter} \cite{Simon2014}. In contrast to the usual bright colors and inwards guiding shape of other flowers, these bat-pollinated plants lure bats by relying on their acoustic reflections, achieved by floral parts shaped specifically to make their echos very conspicuous. We can make bio-inspired sonar landmarks based on different flowers by mimicking these shapes. \\
    One practical example is the dish-shaped leaf of the Cuban liana \textit{Marcgravia evenia}: not only can we approximate the leaf with a spherical reflector, but there is also an opportunity to differentiate the echos of different replicas by adjusting individual properties such as the radius, depth, and edge finish \cite{Simon2020}. In these experiments by Simon et al., an array of ultrasonic transducers was used. \\
    % Continuation
    In this research, we present the possibility of using a real-time imaging sonar with a single transducer to recognize artificial landmarks based on dish leaves of the \textit{Marcgravia evenia} in real-time. The following section details the sensor hardware, landmark design, and algorithm using a support vector machine (SVM) for landmark classification and detection. This is followed by the experimental results for the proposed system,  and finally, this paper is closed off by the conclusions and future work in the final section.
    
    \section{Landmark Detection \& Classification} 

    \begin{figure}
        \centering
        \includegraphics[width=0.75\linewidth]{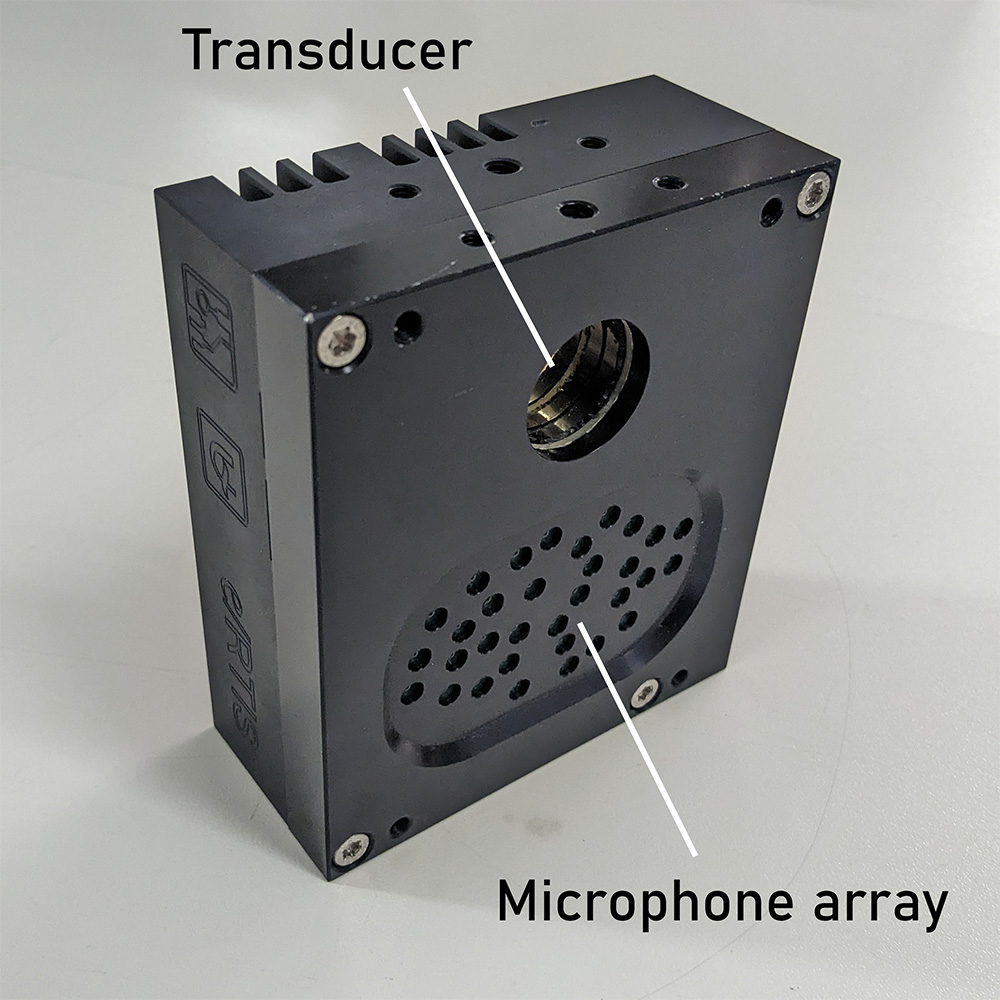}
        \caption{Embedded Real-Time Imaging Sonar.}
        \label{fig:ertis}
    \end{figure}
        
    \subsection{3D Sonar}\label{subsec:sonar}
        The sonar module used is an Embedded Real-Time Imaging Sonar (eRTIS)\cite{Steckel2020} equipped with one SensComp 7000 transducer and 32 Knowles SPH0641LU4H-1 omnidirectional microphones scattered irregularly inside of an ellipsoidal boundary in its frontal face \cite{Kerstens2019}. It also houses an embedded computing platform using an NVIDIA Jetson module system for onboard GPU-accelerated signal processing of the ultrasonic signals \cite{Jansen2019}. \\
        The eRTIS sensor outputs a data stream of pulse-density modulated binary data sequences from the microphones. This is converted to a \SI{450}{\kilo\hertz} sampled digital audio signal onboard the eRTIS module. The resulting audio file is either processed on the eRTIS or stored to be processed offline by the landmark detection algorithm at a later point in time. The sensor is shown in Figure \ref{fig:ertis}. \\ 
        To match the behavior of a bat identifying the \textit{Marcgravia evenia}, the emitted call should match as close as possible. As discussed in \cite{Simon2020}, the ideal chirp would be broadband and be a linear frequency sweep between \SI{180}{\kilo\hertz} to \SI{30}{\kilo\hertz} in \SI{1}{\ms}. To pertain more energy in the signal, a long sweep of \SI{6}{\ms} between \SI{160}{\kilo\hertz} and \SI{10}{\kilo\hertz} can also be used. In practice, the transducer and microphones on the eRTIS cannot produce and record frequencies that high. \\
        Another significant design difference is that the eRTIS sensor only has a single transducer compared to the array of 14 broadband transducers used in the earlier experiments by Simon et Al. \cite{Simon2020}. Each of those transducers would emit sequentially to be able to calculate the landmark directionality, causing the detection algorithm to perform rather slowly. 

    \begin{figure}
        \centering
        \includegraphics[width=\linewidth]{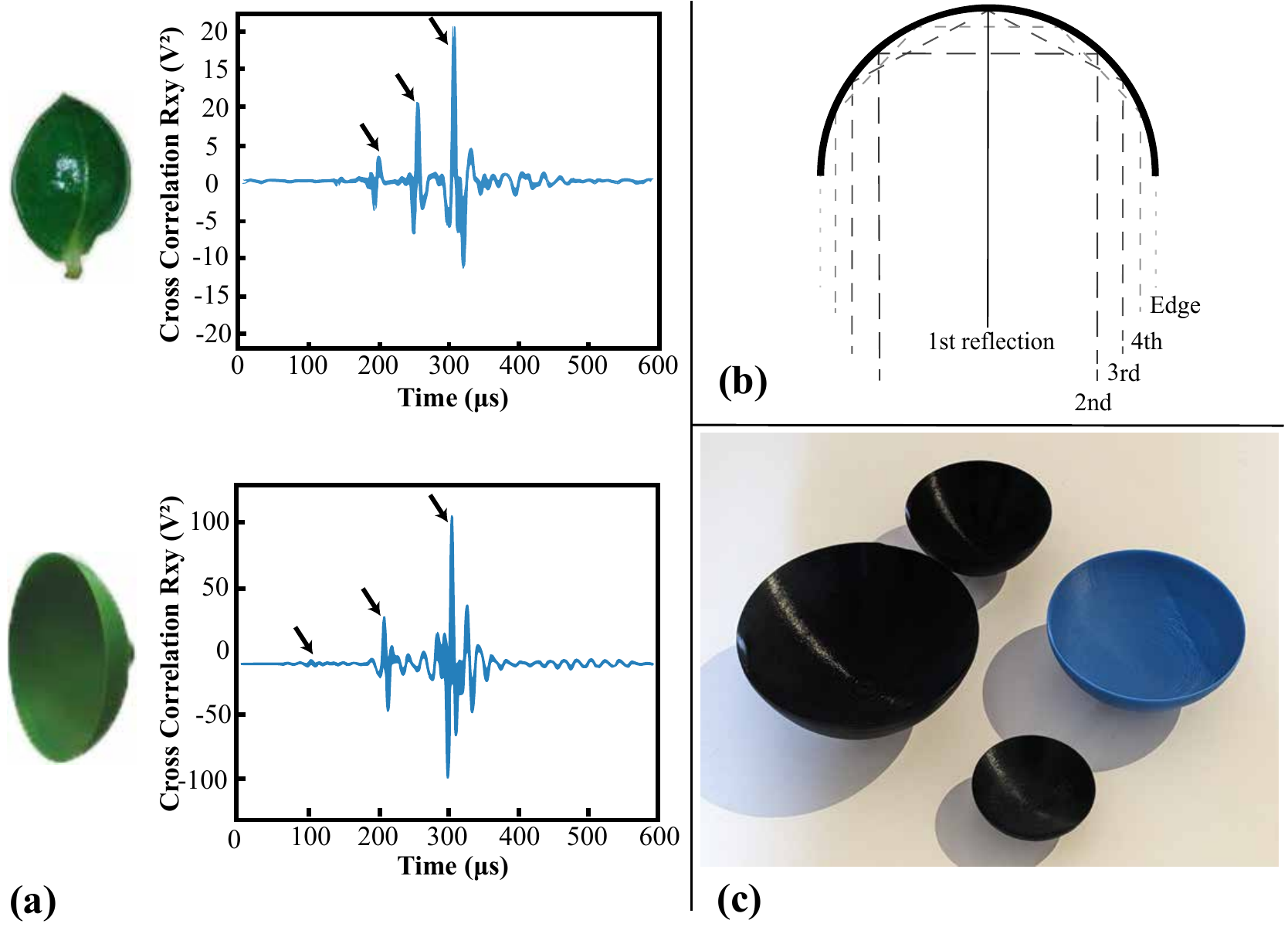}
        \caption{(a) Comparison of a natural (top) and synthetic (bottom) \textit{Marcgravia evenia} leaf showing the impulse response of the echo made from the frontal angle. The arrows indicate the three prominent peaks originating from the edges and the inner concave surface. The synthetic dish shown here has a radius of \SI{35}{\mm} and a depth of \SI{25}{\mm} and uses tapered edges. \cite{Simon2020} (b) The reflection paths illustrated for the spherical shape of the leaf of a \textit{Marcgravia evenia}. (c) 3D printed artificial landmarks with tapered edges and a depth of \SI{30}{\percent} of the total diameter with respective aperture radii of \SI{35}{\mm}, \SI{48}{\mm}, \SI{58}{\mm}, and \SI{70}{\mm} respectively.}
        \label{fig:reflectors}
    \end{figure}
            
    \subsection{Landmarks}
        The landmarks are based on research by Simon et al. \cite{Simon2020}, where different bat-pollinated flowers were examined for their acoustic profile and how bats can recognize them. \\ The \textit{Marcgravia evenia} has developed a unique dish leaf to be recognizable to bats. The leaf has a spherical body that, when struck by sound from the convex side, will reflect acoustic energy to the source through the very center of the dish leaf. Sound will also return after following any secondary paths and through edge reflections. Thus, the resulting echo will consist of a small peak originating from the edge reflections, a prominent peak from the center of the landmark, and subsequent peaks coming from the secondary paths in rapid succession, as shown in Figures \ref{fig:reflectors}a and \ref{fig:reflectors}b. These properties are unique and depend only on the metrics of the landmark, such that we can differentiate differently sized landmarks based on the timings and amplitude of these peaks. Furthermore, they are direction-independent \cite{test}. This forms the basis for the artificial landmark design, consisting of a spherical body. Since we want to maximize the energy in the second and third peaks, we can minimize the edge reflections by tapering the edges. The final artificial landmarks have a depth of \SI{30}{\percent} of their diameter and are 3D printed in polylactic acid (PLA). Some examples are shown in Figure \ref{fig:reflectors}c.

    \subsection{SVM Classification}
        Due to the difference in sonar sensor design as discussed in Subsection \ref{subsec:sonar}, and the goal to reach the real-time operation of the detection algorithm, a different approach had to be used as to the probabilistic algorithm to detect the different peaks as described in \cite{Simon2020}.  
        This algorithm, combined with the eRTIS sensor, caused multiple issues. Most notably, the limited bandwidth of the emitted call possible by the transducer used on the eRTIS causes a lower time resolution due to the Rayleigh criterion. \newpage \noindent Furthermore, since the radii of the different landmarks are pretty close to each other and thus so are the timings between the peaks, the original detection algorithm will sometimes result in positive detection of multiple landmarks with different radii despite only one present. When a landmark is oriented away from the transducer, it can prevent secondary paths from forming. The limiting angle depends on the ratio of the depth of the landmark and the radius of the sphere it is formed around. Finally, if a landmark is too far away or in the side lobe of the transducer, the peaks do not have enough amplitude to stand out from other reflections originating from the main lobe. \\
        Alternatively, we propose to use Support Vector Machines (SVMs) to differentiate between echos originating from different landmarks, as SVMs are widely and successfully used in classifying audio fragments and pattern recognition, among others \cite{LeiChen2006}. Additionally, we inquired if SVMs could differentiate between audio fragments that contain landmark reflections and those that do not. And finally, be more robust against the above-mentioned difficult conditions where the peak detection probabilistic algorithm failed. 

\section{Experimental Results}

    \subsection{SVM Training}
        The training data set consists of recordings made in seven vastly different environments, totaling 944 recordings taken in multiple positions and angles in each environment. They cover four landmarks with respective aperture radii of \SI{35}{\mm}, \SI{48}{\mm}, \SI{58}{\mm}, and \SI{70}{\mm}. The environments range from open lab areas to a packed garage with landmarks between \SI{60}{\cm} and \SI{420}{\cm} from the eRTIS module. The recordings are labeled by landmark size and cropped to the samples that match the range at which the landmark was placed, with a \SI{\pm20}{\cm} margin, leaving us with precisely 1024 samples. After that, we apply a window function to dampen the start and end of the fragment and a band-pass filter between \SI{30}{\kilo\hertz} and \SI{100}{\kilo\hertz}. This bandwidth is divided into 20 linearly spaced frequency bins for training the SVM on the spectrum magnitude obtained through a regular fast Fourier transform. The SVM uses a linear kernel, one-versus-one coding, and is cross-validated 20 times during training.

    \subsection{Validation}
\begin{figure}
        \centering
        \includegraphics[width=0.98\linewidth]{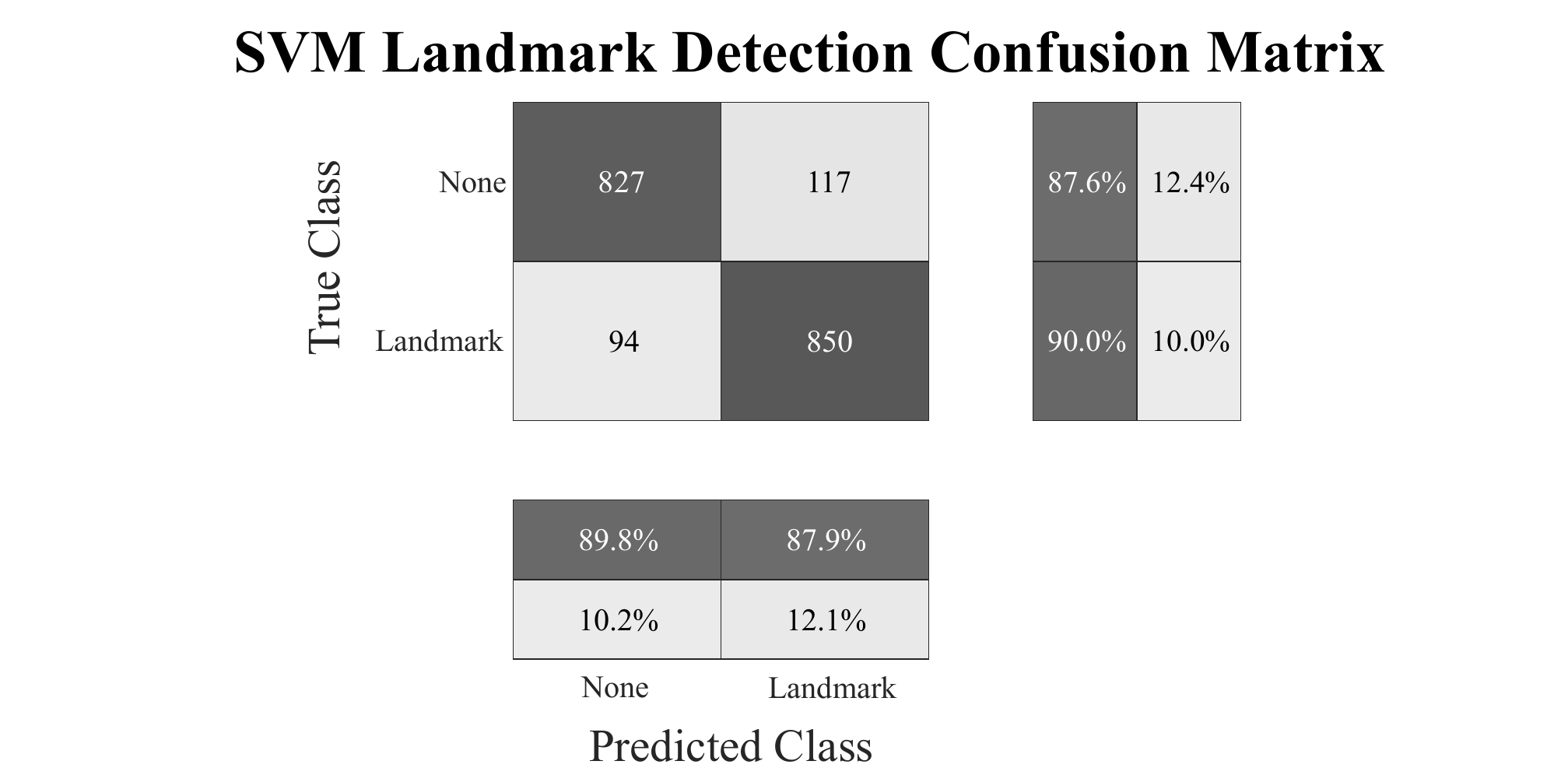}
        \caption{The results of the landmark detection (landmark present or none present) using SVMs in the form of a confusion matrix.}
        \label{fig:results_confusion_detection}
    \end{figure}

\begin{figure}
        \centering
        \includegraphics[width=0.98\linewidth]{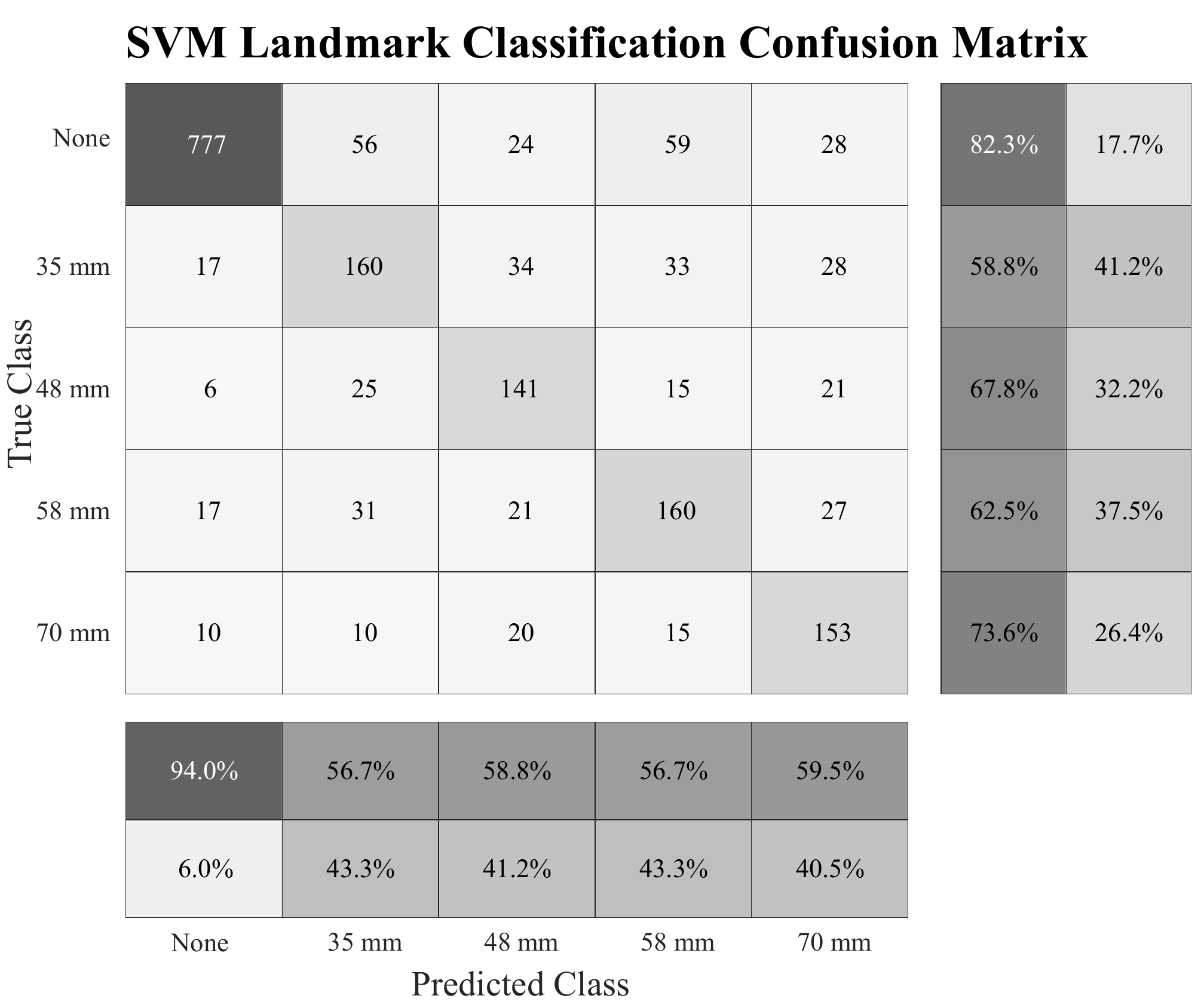}
        \caption{The landmark detection and classification results using SVMs for four different landmark sizes in the form of a confusion matrix.}
        \label{fig:results_confusion_classification}
    \end{figure}

        The resulting SVMs successfully identified whether or not a landmark was present in a fragment with an accuracy of \SI[separate-uncertainty=true]{89.1(0.8)}{\percent}. The confusion matrix of this validation can be seen in Figure \ref{fig:results_confusion_detection}. When differentiating between all four categories, it had a success rate of \SI[separate-uncertainty=true]{67.2(1.3)}{\percent}. The confusion matrix of this validation can be seen in Figure \ref{fig:results_confusion_classification}. The overall accuracy was \SI[separate-uncertainty=true]{73.8(0.9)}{\percent}.
    
\section{Conclusions \& Future Work}

    Using an embedded real-time imaging sonar with a singular transducer with limited bandwidth in both the transducer and receiver array creates a unique constraint to solving the problem of detecting unique acoustic reflectors as landmarks. We proposed using SVMs to identify and classify landmarks in acoustic reflections more reliably as a solution. The experimental results show that it is successful in this task, particularly in landmark detection. The differentiation between the different landmark sizes is less accurate but can be potentially increased by a more extensive training dataset. \\
    Additionally, using the SVM classification method on the frequency spectrum of directional audio fragments can be further explored. Further expansion could create even more distinct acoustic landmarks by using multiple acoustic reflectors in unique patterns similar to fiducial markers for cameras. With regards to obtaining a landmark's bearing, beamforming can be implemented to use the microphone array's properties to find the landmark's direction once the SVM has classified the landmark. Different techniques, such as deep learning-based classifiers (e.g., convolutional neural networks), have also shown promising results and could improve the be more robust than using SVM in the future \cite{test, Kroh2019, Dmitrieva2017}. Finally, a SLAM algorithm can implement the landmark detection system to provide more stable loop closure events.

\clearpage

\bibliographystyle{IEEEtran}
\bibliography{ms}

\begin{thebibliography}{10}
\providecommand{\url}[1]{#1}
\csname url@samestyle\endcsname
\providecommand{\newblock}{\relax}
\providecommand{\bibinfo}[2]{#2}
\providecommand{\BIBentrySTDinterwordspacing}{\spaceskip=0pt\relax}
\providecommand{\BIBentryALTinterwordstretchfactor}{4}
\providecommand{\BIBentryALTinterwordspacing}{\spaceskip=\fontdimen2\font plus
\BIBentryALTinterwordstretchfactor\fontdimen3\font minus
  \fontdimen4\font\relax}
\providecommand{\BIBforeignlanguage}[2]{{%
\expandafter\ifx\csname l@#1\endcsname\relax
\typeout{** WARNING: IEEEtran.bst: No hyphenation pattern has been}%
\typeout{** loaded for the language `#1'. Using the pattern for}%
\typeout{** the default language instead.}%
\else
\language=\csname l@#1\endcsname
\fi
#2}}
\providecommand{\BIBdecl}{\relax}
\BIBdecl

\bibitem{Esrafilian2017}
O.~Esrafilian and H.~D. Taghirad, ``Autonomous flight and obstacle avoidance of
  a quadrotor by monocular slam,'' \emph{4th RSI International Conference on
  Robotics and Mechatronics, ICRoM 2016}, pp. 240--245, 3 2017.

\bibitem{Bescos2021}
B.~Bescos, C.~Campos, J.~D. Tardos, and J.~Neira, ``Dynaslam ii:
  Tightly-coupled multi-object tracking and slam,'' \emph{IEEE Robotics and
  Automation Letters}, vol.~6, pp. 5191--5198, 7 2021.

\bibitem{Debeunne2020}
C.~Debeunne and D.~Vivet, ``A review of visual-lidar fusion based simultaneous
  localization and mapping,'' \emph{Sensors 2020, Vol. 20, Page 2068}, vol.~20,
  p. 2068, 4 2020.

\bibitem{Batslam}
J.~Steckel and H.~Peremans, ``Batslam: Simultaneous localization and mapping
  using biomimetic sonar,'' \emph{PLOS ONE}, vol.~8, no.~1, pp. 1--11, 01 2013.

\bibitem{Chen2006}
J.~Benesty, J.~Chen, Y.~Huang, and J.~Dmochowski, ``On microphone-array
  beamforming from a mimo acoustic signal processing perspective,'' \emph{IEEE
  Transactions on Audio, Speech, and Language Processing}, vol.~15, no.~3, pp.
  1053--1065, 2007.

\bibitem{michel2006history}
U.~Michel, ``History of acoustic beamforming,'' in \emph{1st. Berlin
  Beamforming Conference}, 2006.

\bibitem{thrun2005}
S.~Thrun, W.~Burgard, and D.~Fox, \emph{Probabilistic robotics}.\hskip 1em plus
  0.5em minus 0.4em\relax Cambridge, Mass.: MIT Press, 2005.

\bibitem{Sefati2017}
M.~Sefati, M.~Daum, B.~Sondermann, K.~D. Kreisköther, and A.~Kampker,
  ``Improving vehicle localization using semantic and pole-like landmarks,'' in
  \emph{2017 IEEE Intelligent Vehicles Symposium (IV)}, 2017, pp. 13--19.

\bibitem{Grisetti2010}
G.~Grisetti, R.~Kummerle, C.~Stachniss, and W.~Burgard, ``A tutorial on
  graph-based slam,'' \emph{IEEE Intelligent Transportation Systems Magazine},
  vol.~2, pp. 31--43, 12 2010.

\bibitem{kalaitzakis2021fiducial}
M.~Kalaitzakis, B.~Cain, S.~Carroll, A.~Ambrosi, C.~Whitehead, and
  N.~Vitzilaios, ``Fiducial markers for pose estimation: Overview, applications
  and experimental comparison of the artag, apriltag, aruco and stag markers,''
  \emph{Journal of Intelligent \& Robotic Systems}, vol. 101, pp. 1--26, 2021.

\bibitem{test}
R.~Simon, K.~Bakunowski, A.~E. Reyes-Vasques, M.~Tschapka, M.~Knörnschild,
  J.~Steckel, and D.~Stowell, ``Acoustic traits of bat-pollinated flowers
  compared to flowers of other pollination syndromes and their echo-based
  classification using convolutional neural networks,'' \emph{PLOS
  Computational Biology}, vol.~17, p. e1009706, 12 2021.

\bibitem{Simon2014}
R.~Simon, M.~Knörnschild, M.~Tschapka, A.~Schneider, N.~Passauer, E.~K. Kalko,
  and O.~von Helversen, ``Biosonar resolving power: Echo-acoustic perception of
  surface structures in the submillimeter range,'' \emph{Frontiers in
  Physiology}, vol. 5 FEB, p.~64, 2 2014.

\bibitem{Simon2020}
R.~Simon, S.~Rupitsch, M.~Baumann, H.~Wu, H.~Peremans, and J.~Steckel,
  ``Bioinspired sonar reflectors as guiding beacons for autonomous
  navigation,'' \emph{Proceedings of the National Academy of Sciences of the
  United States of America}, vol. 117, pp. 1367--1374, 1 2020.

\bibitem{Steckel2020}
\BIBentryALTinterwordspacing
J.~Steckel, ``{3dsonar.eu},'' 2020. [Online]. Available: \url{www.3dsonar.eu}
\BIBentrySTDinterwordspacing

\bibitem{Kerstens2019}
R.~Kerstens, D.~Laurijssen, and J.~Steckel, ``Ertis: A fully embedded real time
  3d imaging sonar sensor for robotic applications,'' \emph{Proceedings - IEEE
  International Conference on Robotics and Automation}, vol. 2019-May, pp.
  1438--1443, 5 2019.

\bibitem{Jansen2019}
W.~Jansen, D.~Laurijssen, R.~Kerstens, W.~Daems, and J.~Steckel, ``{In-Air
  Imaging Sonar Sensor Network with Real-Time Processing Using GPUs},'' in
  \emph{3PGCIC 2019: Advances on P2P, Parallel, Grid, Cloud and Internet
  Computing}, vol.~96.\hskip 1em plus 0.5em minus 0.4em\relax Springer, 2020,
  pp. 716--725.

\bibitem{LeiChen2006}
L.~Chen, S.~Gündüz, and M.~T. Özsu, ``Mixed type audio classification with
  support vector machine,'' \emph{2006 IEEE International Conference on
  Multimedia and Expo, ICME 2006 - Proceedings}, vol. 2006, pp. 781--784, 2006.

\bibitem{Kroh2019}
P.~K. Kroh, R.~Simon, and S.~J. Rupitsch, ``Classification of sonar targets in
  air: A neural network approach,'' \emph{Sensors}, vol.~19, no.~5, 2019.

\bibitem{Dmitrieva2017}
M.~Dmitrieva, M.~Valdenegro-Toro, K.~Brown, G.~Heald, and D.~Lane, ``Object
  classification with convolution neural network based on the time-frequency
  representation of their echo,'' in \emph{2017 IEEE 27th International
  Workshop on Machine Learning for Signal Processing (MLSP)}, 2017, pp. 1--6.

\end{thebibliography}

\end{document}